\documentclass[10pt, a4paper]{article}
\usepackage{lrec2022} 
\usepackage{multibib}
\newcites{languageresource}{Language Resources}
\usepackage{graphicx}
\usepackage{multirow,tabularx}   
\usepackage{soul}

\usepackage{titlesec}
\titleformat{\section}{\normalfont\large\bfseries\center}{\thesection.}{1em}{}
\titleformat{\subsection}{\normalfont\SmallTitleFont\bfseries\raggedright}{\thesubsection.}{1em}{}
\titleformat{\subsubsection}{\normalfont\normalsize\bfseries\raggedright}{\thesubsubsection.}{1em}{}
\renewcommand\thesection{\arabic{section}}
\renewcommand\thesubsection{\thesection.\arabic{subsection}}
\renewcommand\thesubsubsection{\thesubsection.\arabic{subsubsection}}

\usepackage{epstopdf}
\usepackage[utf8]{inputenc}

\usepackage{hyperref}
\usepackage{xstring}

\usepackage{color}

\title{PerCQA: Persian Community Question Answering Dataset}

\name{Naghme Jamali, Yadollah Yaghoobzadeh, Hesham Faili} 

\address{School of Computer Science, Institute for Research in Fundamental Sciences\\ School of Electrical and Computer Engineering, College of Engineering, University of Tehran\\ School of Electrical and Computer Engineering, College of Engineering, University of Tehran\\
         Tehran, Iran \\
         naghme.jamali@ipm.ir, y.yaghoobzadeh@ut.ac.ir, hfaili@ut.ac.ir}

\abstract{
\begin{minipage}{\textwidth}
Community Question Answering (CQA) forums provide answers for many real-life questions. Thanks to the large size, these forums are very popular among machine learning researchers. Automatic answer selection, answer ranking, question retrieval, expert finding and fact-checking are example learning tasks performed using CQA data. 
In this paper, we present \textbf{PerCQA}, the ﬁrst Persian dataset for CQA. This dataset contains the questions and answers crawled from the most well-known Persian forum. After data acquisition, we provide rigorous annotation guidelines in an iterative process, and then the annotation of question-answer pairs in SemEvalCQA format. PerCQA contains 989 questions and 21,915 annotated answers. We make PerCQA publicly available to encourage more research in Persian CQA. We also build strong benchmarks for the task of answer selection in PerCQA by using mono- and multi-lingual pre-trained language models.
\end{minipage}
 \\ \newline  
}
\begin{document}

\maketitleabstract

\section{Introduction}
A Community Question Answering (CQA) platform allows web users to get answers for their questions from domain experts. With the advent of websites such as Yahoo! Answers\footnote{\url{https://answers.yahoo.com/}}, Stack Exchange\footnote{\url{https://stackexchange.com/.}} and Quora\footnote{\url{https://www.quora.com/.}}, CQA has attracted a lot of attention.
CQA is considered as a reliable source for acquiring required knowledge to solve problems that cannot be solved directly by searching on web pages. Due to the great expansion of these forums in terms of topics and number of users in different languages, manual checking and verification of contents by the administrators is very challenging. Therefore, automatic solutions are required especially to direct users to the most appropriate answers for each question. 

The ability to automatically find relevant questions to reuse their existing answers (i.e., question retrieval) and searching for relevant answers among many responses for a given question (i.e., answer selection) are famous tasks in CQA. Most existing work in this area is conducted using publicly available datasets, such as SemEvalCQA~\cite{nakov-etal-2015-semeval,nakov-etal-2016-semeval-2016,nakov-etal-2017-semeval}, TREC QA~\cite{wang2007what}, Wiki QA~\cite{yang2015wikiqa}, and InsuranceQA~\cite{feng2015applying}. Here, we focus on datasets in Persian language. One significant limitation for developing automatic systems to manage the Persian CQA forums is the lack of a such datasets in Persian language. This affects the quality and usability of these forums as well. By doing research on these forums, their quality can be improved because the users are able to get the answers they desire more efficiently.

In this study, we build a dataset for Persian CQA, called \emph{PerCQA}. It consists of questions and answers posted by the users of the most popular Persian forum, named Ninisite\footnote{\url{www.ninisite.com}}. Our main focus is on data gathering, preparation, and annotation. A web scraping tool is developed for extracting useful information from Ninisite. The most important step in preparing a high-quality dataset is the development of a detailed and consistent annotation guideline. Our annotation guideline affects the agreement between annotators significantly. 
We also build an annotation tool to reduce the labeling time.

Data annotation in PerCQA consists of two stages: question annotation for selecting appropriate questions and answer annotation with three labels (``Good'', ``Bad'', ``Potential''). As a result, \emph{PerCQA} is built with 989 questions and 21,915 answers. This dataset is structurally similar to SemEval 2015 English CQA dataset~\cite{nakov-etal-2015-semeval}, which has 3,229 questions and 20,162 answers. 

Furthermore, in order to evaluate PerCQA and analyze the answer selection task, several extensive experiments are performed with non-contextualized, namely \emph{Word2vec}~\cite{DBLP:journals/corr/abs-1301-3781}, \emph{fastText}~\cite{bojanowski2017enriching} and contextualized word embeddings learnd by pretrained language models (PLMs), namely \emph{BERT}~\cite{devlin-etal-2019-bert}, \emph{ParsBERT}~\cite{farahani2021parsbert}. We further improve our results by transferring knowledge from English datasets using multilingual PLMs such as \emph{mBERT}~\cite{devlin-etal-2019-bert} and XLM-RoBERTa (\emph{XLM-R})~\cite{conneau-etal-2020-unsupervised}.

We use \emph{PV-Cnt}~\cite{yang2015wikiqa}, \emph{BiLSTM-attention}~\cite{tan2015lstm}, \emph{RCNN}~\cite{zhou2018recurrent} and \emph{CETE}~\cite{laskar2020contextualized} as our baseline systems that employ word embeddings. Our experimental results demonstrate that ParsBERT and XLM-R embeddings using by CETE system outperform all other baselines. We find that XLM-R overtakes ParsBERT by up to about +3\% macro F1-score on PerCQA by transferring knowledge from SemEvalCQA English datasets. 

Our main contributions can be summarized as the followings: 
\begin{itemize}
    \item We build and release PerCQA\footnote{\url{https://github.com/PerCQA}}, the first Persian dataset for CQA, to enhance the research and applications of CQA tasks in Persian.
    \item We apply several state-of-the-art methods to our dataset and set strong baselines for the task of answer selection in PerCQA.
\end{itemize}

In the next section, some previous research and existing CQA datasets are reviewed. Section \ref{Section3} describes the process of creating our dataset in detail. The structure of PerCQA and its statistics is presented in Section \ref{Section4}. In Section \ref{Section5} experimental results and analysis are presented. Finally, the Section \ref{Section6} is the concluding remarks.   


\section{Related Work}

In this section, we look at existing CQA datasets that are widely used for evaluating the CQA tasks and introduce some of the tasks in these forums and describe some various models in answer selection task.
\begin{table}[!h]
\begin{center}
    \begin{tabular}{c|c|c|c}
    \hline
    Dataset                        & Set   & \#Questions & \#Answers \\ \hline
    \multirow{3}{*}{SemEval 2015}  & Train & 2600        & 16541     \\
                               & Dev   & 300         & 1654      \\
                               & Test  & 329         & 1976      \\ \hline
    \multirow{3}{*}{SemEval 2016}  & Train & 4879        & 36198     \\
                               & Dev   & 244         & 2440      \\
                               & Test  & 327         & 3270      \\ \hline
    \multirow{3}{*}{SemEval 2017}  & Train & 4879        & 36198     \\
                               & Dev   & 244         & 2440      \\
                               & Test  & 293         & 2930      \\ \hline
    \multirow{3}{*}{Insurance QA}  & Train & 12889       & 21325     \\
                               & Dev   & 2000        & 3354      \\
                               & Test  & 2000        & 3308      \\ \hline
    \multirow{3}{*}{Yahoo CQA}     & Train & 50112       & 253440    \\
                               & Dev   & 6289        & 31680     \\
                               & Test  & 6283        & 31680     \\ \hline
    \multirow{3}{*}{WikiQA (RAW)}  & Train & 2118        & 20360     \\
                               & Dev   & 296         & 2733      \\
                               & Test  & 633         & 6165      \\ \hline
    \multirow{3}{*}{TREC-QA (RAW)} & Train & 1229        & 53417     \\
                               & Dev   & 82          & 1148      \\
                               & Test  & 100         & 1517      \\ \hline
    \end{tabular}
    \caption{Statistics of various CQA datasets.}
    \label{table:Statistics of various DataSets}
\end{center}
\end{table}

\subsection{CQA Datasets}
There are many CQA datasets released in different languages to date. The following datasets are available in English: SemEval~\cite{nakov-etal-2015-semeval}~\cite{nakov-etal-2017-semeval}, TREC QA~\cite{wang2007what}, WikiQA~\cite{yang2015wikiqa}, Insurance QA~\cite{feng2015applying}, and Yahoo! Answers~\cite{qiu2015convolutional}. There are three versions of the SemEval dataset (2015, 2016, and 2017), crawled from the Qatar Living forum which each question has attributes such as question category, question type, and question date. The TREC-QA dataset, provided by TREC-QA track 8-13, is the most widely used benchmark for testing QA and CQA models. Another popular dataset for evaluating answer selection systems is WikiQA. This dataset is collected from Bing query logs. The Insurance QA dataset is a non-factoid QA dataset from the insurance domain. The Yahoo! Answers dataset was generated by~\cite{qiu2015convolutional}, using the Computer and Internet category is resolved questions in Yahoo! Answers. 

Other datasets from CQA are also released in other languages like \emph{Arabic} or \emph{Chinese} in addition to English. For Arabic, ``SemEval-2015 CQA-subtask A'' provided a dataset that was crawled from the \emph{Fatwa}\footnote{\url{http://fatwa.islamweb.net/}} website for its answer selection shared task~\cite{nakov-etal-2015-semeval} and also in~\cite{nakov-etal-2017-semeval} ``SemEval-2017 CQA-subtask D'' only was used one of the existing websites,namely Altibbi\footnote{\url{http://www.altibbi.com}}. JEC-QA~\cite{zhong2020jec} is a Chinese question answering dataset of Chinese law forums for legal advice that containing a large amount of legal knowledge. It collected from the National Judicial Examination of China and is available from\footnote{\url{https://jecqa.thunlp.org/}}. Table \ref{table:Statistics of various DataSets} presents the statistics of some datasets in CQA and according to the research we have done, there is no Persian dataset in CQA.

\subsection{CQA Tasks}
Following the creation of datasets in different languages, different types of research have been conducted on CQA platforms. ``Question similarity'' task in CQA forums is to retrieve a collection of questions similar to the question that the user has asked. In SemEval-2016/2017 (task3-Subtask B)~\cite{nakov-etal-2017-semeval}~\cite{nakov-etal-2016-semeval-2016}, ``Question-Question Similarity'' task has been included as a benchmark task. \cite{kunneman-etal-2019-question} demonstrated adjusting preprocessing and word similarity settings improved the result of identifying duplicate questions. The goal of ``Question Retrieval'' in CQA is to find existing and semantically equivalent questions. Answers to the queried questions are derived from the best answers to these similar questions. Numerous studies have been conducted in this field~\cite{zhou2015learning},~\cite{othman2017a},~\cite{othman2019manhattan},~\cite{10.1145/3397271.3401143}. 

It is very important and useful to find users who have expertise to answer your questions. Therefore, ``Expert Finding'' is one of the important tasks in CQA. To this aim,~\cite{ghasemi2021user}'s model, for extracting users’ embeddings, applied node2vec~\cite{grover2016node2vec} and matrix factorization-based embedding~\cite{qiu2018network}. In ``Answer Selection'' task, the answers provided for each question are classified to determine relevant and non-relevant answers.  In answer ranking task, semantic similarity is used to rank the answers relevant to a question. This field is covered in the works of~\cite{mihaylov2016semanticz},~\cite{nakov2016it}, and~\cite{omari2016novelty}. 

\subsection{Answer Selection Models}
There are basically two main types of answer classification methods: feature-based methods and deep-learning methods. Several early works used feature-based methods for explicitly modeling the semantic relation between the question and answer~\cite{nakov-etal-2015-semeval},~\cite{huang2007extracting},~\cite{agichtein2008finding},~\cite{nicosia2015qcri}. JAIST~\cite{tran-etal-2015-jaist}, HITSZ-ICRC~\cite{hou-etal-2015-hitsz} and QCRI~\cite{nicosia2015qcri} utilize typical features such as special component features, word matching features, non-textual features and topic-modeling-based features. A simple classifier such as Support Vector Machine (SVM) or KNN is applied to the features and easily were selected. In~\cite{yang2015wikiqa}, a lexical-semantic feature method employing word/lemma matching is considered for baseline. 

The use of deep learning based methods reduces feature engineering to a large extent, as they automatically learn all features through end-to-end training. In light of great advancements in deep learning neural networks, considerable recent researches have applied deep learning-based methods to perform answer classification in CQA~\cite{tan2015lstm},~\cite{xiang2017answer},~\cite{xiang2016incorporating},~\cite{wen2019joint},~\cite{yang2019advanced}. Typically, they~\cite{mihaylov2016semanticz},~\cite{nakov2016it},~\cite{omari2016novelty},~\cite{zhou2015learning},~\cite{othman2019manhattan} use a Convolutional Neural Network (CNN) or Long Short Term Memory (LSTM) network for matching the question and answer. \emph{Word2vec}~\cite{DBLP:journals/corr/abs-1301-3781}, \emph{Glove}~\cite{pennington-etal-2014-glove} and \emph{fastText}~\cite{joulin2016fasttext} as non-contextualized word embedding provide fixed representation for each word and do not capture its context in different sentences. 

Recently, for learning sentence representation, various attention models based on the transformer model have been proposed~\cite{vaswani2017attention}. Transformer networks also serve as an encoder or decoder for some models in different tasks~\cite{DBLP:journals/corr/abs-1803-11175},~\cite{Radford2018ImprovingLU}. Nowadays, \emph{BERT}~\cite{devlin-etal-2019-bert} and \emph{RoBERTA}~\cite{liu2019roberta} are used widely as contextualized word embeddings. In~\cite{laskar2020contextualized}, a model is presented which integrates contextualized embeddings with the transformer encoder (CETE) for sentence similarity modeling. CETE is based on contextualized embeddings (BERT, RoBERTA and ELMo (Embeddings from Language Models~\cite{peters2018deep})). There are two approaches in CETE, namely, features-based and finetuning-based. We use the feature-base approach here.

Most of these models are geared towards English, leaving multilingual models with limited resources to cover other languages. Multilingual Language Models such as mBERT~\cite{devlin-etal-2019-bert}, XLM~\cite{conneau2019cross}, and XLM-R~\cite{conneau-etal-2020-unsupervised}, apply the power of pretraining to multiple languages. We use the standard zero-shot and cross-lingual transfer learning such as mBERT and XLM-RoBERTa (XLM-R) to achieve more accurate representations for our task. We also employ ParsBERT~\cite{farahani2021parsbert}, the Persian pretrained language model, and compare it 
with the multi-lingual representation models. 
We utilize them as  encoders and measure the similarity between the vector representation of two sentences produced by these models. 
\begin{table*}[ht]
\begin{center}
    \begin{tabular}{l|l}
    \hline
    \multirow{2}{*}{1. Data acquisition} & a. Exploring various CQA Persian websites.                                                   \\
                                     & b. Crawling the desired website(www.ninisite.com).                                           \\ \hline
    \multirow{6}{*}{2. Data annotation}  & a. Setting Annotation Guidelines(AG) for the questions(Labels: Valid, Invalid).              \\
                                     & b. Setting AG for the answers(labels: Good, Bad, Potential).                                 \\
                                     & c. Developing an application for manual labeling.                                            \\
                                     & d. Crowdsourcing: question labeling(4people), answer labeling(12people).                     \\
                                     & e. Evaluation of labeling(by Cohen's Kappa coefficient)(4people).                            \\
                                     & f. Repeating the previous steps with improving the quality of AG and labeling answers again. \\ \hline
    \end{tabular}
    \caption{The steps of construction of PerCQA.}
    \label{table:steps of construction}
\end{center}
\end{table*}

\section{Dataset Construction}
\label{Section3}
A CQA is a powerful mechanism that usually consists of two steps: (i) creating a free-form question (ii) posting various answers, thus creating an extensive collection for the desired question by users in order to obtain specific answers to their questions. The new question can be related to one or more previous questions in the forum. The best answer is sometimes marked in the question-answer threads that the lists of replies are sorted chronologically. The meta-information includes the date of posting, the user who asked/answered, a category question, and answer tags. 

Several research CQA datasets are conducted in English as well as few other languages. We build the first CQA dataset in Persian, called PerCQA, designed for research purposes. To construct the dataset, we initially study the standard CQA datasets in English. We follow the standard approach in CQA datasets and crawl user forums to construct our dataset. Table \ref{table:steps of construction} illustrates a summary of the steps involved in building PerCQA.
In the following subsections, we describe in detail the process of creating PerCQA dataset.

\subsection{Data Acquisition}
We perform a complete analysis of various Persian forums and examine the ranking websites in Alexa Internet \footnote{\url{https://www.alexa.com/siteinfo/ninisite.com}} and Similarweb.com\footnote{\url{https://www.similarweb.com/website/ninisite.com}}. A website was chosen for its question source based on the following characteristics: (i) having the highest rank in Iran, (ii) having numerous users, (iii) containing irrelevant answers in the sequence of replies to a question, (iv) having more than \%80 traffic from search in alexa.com, (v) it is being used by ordinary people, not necessarily specialists in different fields. 

In alexa.com, Ninisite is ranked approximately 20th in Iran and about 2,049 in the world and also it has all the above characteristics. Therefore, \url{www.ninisite.com} is used as our question source and from May 5th, 2017 until December 21st, 2020, about 1400 questions and the corresponding logs are crawled. However, we do not use all of the questions, mainly because some of them are advertisements or polls. Therefore, we need to evaluate the types of questions first to filter out some of them. The description of \emph{Ninisite}'s structure is given in the next subsection.

\subsection{Structure of Ninisite}
Ninisie is the first and largest online forum in Persian for children, mothers and families. This website includes diverse sections. There is a section called ``/tba:dol næzær/'' (idea exchange) with 20 categories. Furthermore, there are specific forums called ``/ta:la:r/'' and any question asked in each forum is called ``/ta:pik/''. Webmasters with this slogan invite their users to participate in the forum. ``Please participate in the Ninesite exchange and connect with thousands of other mothers''. We examine the frequently asked questions about women, children, and medical subjects.

\subsection{Data Annotation}
Data annotation is necessary for this dataset because we aim to run strong supervised learning methods. It is the most important component of our work. There are three steps which comprise question annotation, answer annotation, and evaluation of labeling quality. We develop a tool for data annotation which selects appropriate questions and answers. This tool drastically speeds up collecting datasets and also is an efficient way to collaborate on annotation projects. We crowdsource the annotation through a special platform with 20 users.

\paragraph{Question annotation}
We review about a hundred questions and set up an annotation guideline for them. Table \ref{table:AG for Q} shows annotation guidelines for the questions. For example, ``Hiring a hairdresser, if you want to learn hairdressing, send me a message and I will tell you the conditions.'' or introducing a book with this post, ``This topic makes your life 180 degrees better'' \footnote{\url{https://www.ninisite.com/discussion/topic/7764925}} are marked invalid questions, and it is not included in the dataset because there is no helpful answer to this question.\\
\begin{table}[!h]
\begin{center}
    \begin{tabular}{lc}
    \hline
    \textbf{Type of questions} & \textbf{Label} \\ 
    \hline
    Advertisement              & Invalid \\ 
    \hline
    Polls                      & Invalid \\ 
    \hline
    News                       & Invalid \\ 
    \hline
    \begin{tabular}[l]{@{}l@{}}Less than 3 answers and more\\ than 300 answers
    \end{tabular}              & Invalid \\ 
    \hline
    Collaboration announcement & Invalid \\ 
    \hline
    Otherwise                  & Valid    \\ 
    \hline
    \end{tabular}
\caption{“Annotation guidelines” for questions.}
\label{table:AG for Q}
\end{center}
\end{table}
In our data annotation tool, three stages of the cascaded UI are designed. The first step displays the questions. ``Valid'' and ``Invalid'' labels are assigned to the questions in the question selection method. Valid questions are entered into the database and transferred to the next labeling phase, and invalid ones are deleted. Then, the UI moves on to the next step. After choosing appropriate questions, the system enters into the second stage, labeling the answers.

\paragraph{Answer annotation}
\begin{table}[!h]
\begin{center}
    \begin{tabular}{l|c}
    \hline
    \textbf{Type of answers}           & \textbf{Label} \\ 
    \hline
    \begin{tabular}[c]{@{}l@{}}Dialogue, Advertisement, Not Persian,\\Greeting, Sympathy, stickers,\\ Acknowledgment, Persian typed\\in English, and other comments.
    \end{tabular} & \textit{Bad} \\ 
    \hline
    \begin{tabular}[c]{@{}l@{}}Referred to other sources such as a\\related link or site (URLs) or a special\\page in social media and so on.
    \end{tabular} & \textit{Potential} \\ 
    \hline
    Partial or complete relevant answer to\\the given question & \textit{Good}      \\ 
    \hline
    \end{tabular}
\caption{“Annotation guidelines” for answers.} 
\label{table:AG for A}
\end{center}
\end{table}
\begin{table*}[ht]
\begin{center}
    \begin{tabular}{l}
    \hline
    \textbf{\begin{tabular}[c]{@{}l@{}}Question: What is your suggestion to heal Cervical Disc?  It bothers me a lot. (Valid Question)
    \end{tabular}} \\ 
    \hline
    \begin{tabular}[c]{@{}l@{}}Answer 1: Dr. Mohammad Kamali is a professional Surgeon who can treat you without surgery. I had a rigid\\ Cervical Disc with a lot of pain. After some therapy sessions, his pain disappeared (Good Answer)
    \end{tabular} \\ 
    \hline
    \begin{tabular}[c]{@{}l@{}}Answer 2: I had a cervical disc you should take a careful care and never bend your neck. (Good Answer)
    \end{tabular}\\ 
    \hline
    \begin{tabular}[c]{@{}l@{}}Answer 3: For more information about Cervical Disc, check Dr. Samadian's website and type your questions \\https://drsamadian.com/cervical-disc/ (Potential Answer)\end{tabular}\\ 
    \hline
    Answer 4: What   medicines have been prescribed? (Bad Answer)\\
    \hline
    \begin{tabular}[c]{@{}l@{}}Answer5: The doctor told my mother that her Cervical Disc is in a lousy condition. The nerves are torn.\\ She should pass 15 sessions of physiotherapy, but my mother can't move at all. (Bad Answers)
    \end{tabular}\\ 
    \hline
    \end{tabular}
\caption{The translation of a question and some of its answers in PerCQA.}
\label{table:Sample question and its answers}
\end{center}
\end{table*}
After an internal labeling of a trial dataset (the 100 selected questions) by several independent annotators, we prepare detailed annotation guidelines for labeling answers. The answers are classified as ``Good'', ``Bad'', and ``Potential''. ``Good'' labels are assigned to relevant answers and potentially useful answers are labeled ``Potential''. The irrelevant answers (bad, dialog, non-English, other) take ``Bad" labels. Table \ref{table:AG for A} shows ``Annotation guidelines'' for answers, a type of labeling guide to increase agreement between the annotators. In this phase, twelve annotators tag the answers using their unique user IDs. We ask three workers to tag each answer and select the correct label by majority voting.

\paragraph{Quality assessment}
Assessing the labeling quality of answers is the third stage of our data annotation. We used Cohen Kappa criteria to evaluate the quality. Consequently, 45.28\% of the total data are re-tagged for analysis by two groups consisting of two individuals. The Cohen’s kappa (percentage of agreement) is 80\%, More details are available in Appendix (Data labeling quality).

\section{PerCQA Dataset}
\label{Section4}
There are in total 989 questions and 21,915 corresponding answers in PerCQA. The content of questions and answers is kept mostly unchanged. In this section, the structure of the dataset, its features and statistics are explained.

\subsection{Structure}
The translation of a sample question and a subset of its answers in PerCQA are illustrated in Table \ref{table:Sample question and its answers}. The original version of this question in PerCQA is given in Appendix (Details about PerCQA). 

It is a sequence of questions with the following attributions:
(i) question identiﬁer (QID);
(ii) the questioner Username (QUsername);
(iii) the day and date the question is posted on the forum (QDate);
(iv) the questioners consider a subject for their question and write it (QSubject);
(v) the whole question (QBody). 

Each question is followed by a list of comments and each answer contains the following attributes:
(i) comment identiﬁer (CID);
(ii) identiﬁer of the user posting the comment (CUserID);
(iii) Username of the respondent (CUsername);
(iv) comment body (CBody);
(v) an annotator rating of whether the comment is ``Good'', ``Bad'' and ``Potential'' (CGOLD).

At testing time, CGOLD is hidden, and the system is instructed to predict CGOLD. It can be seen that questions and answers in PerCQA are informal and can be considered as a problem in building a suitable language model. Other challenges include incorrect capitalization and punctuation, misspelling, as well as slang and elongations. To learn word embeddings, we collect unlabeled texts, including 10000 questions and its comments that is posted in the site. 

\subsection{Statistics}
\begin{table*}[ht]
\begin{center}
\begin{tabular}{l|c|c|c|c|c|c}
& \begin{tabular}[c]{@{}c@{}}Number of\\ Questions\end{tabular} & \begin{tabular}[c]{@{}c@{}}Number of \\ Answers\end{tabular} & \begin{tabular}[c]{@{}c@{}}Mean length \\ of questions\end{tabular} & \begin{tabular}[c]{@{}c@{}}Mean length\\ of Answers\end{tabular} & \begin{tabular}[c]{@{}c@{}}Median  length\\ of questions\end{tabular} & \begin{tabular}[c]{@{}c@{}}Median length\\ of answers\end{tabular} \\ 
\hline 
\textbf{Train} (70\%) & 692 & 15,454 & 171 & 84 & 129 & 61\\ 
\hline
\textbf{Dev} (10\%)   & 99  & 2,164  & 174 & 92 & 128 & 64 \\ 
\hline
\textbf{Test} (20\%)  & 198 & 4,297  & 176 & 93 & 130 & 64 \\ 
\hline
\textbf{Total}       & 989 & 21,915 & 173 & 89 & 129 & 62 \\ 
\end{tabular}
\caption{PerCQA statistics.}
\label{table:Statistics of PerCQA}
\end{center}
\end{table*}
The statistics of the dataset are given in Table \ref{table:Statistics of PerCQA}. Since one of the extremely important features  in feature-based methods is ``the length of questions and answers'', its mean and median are reported in this table for a more comprehensive review. The existence of one-word questions and answers, as well as very long ones in the dataset, motivate us to report the median in addition to the mean.
\begin{table*}[ht]
\begin{center}
    \begin{tabular}{l|l}
    \hline
    \textbf{Category} & \textbf{Features} \\ 
    \hline
    \textbf{Question Specification} & Number of characters in each question(Length of the question)\\
    \hline
    \textbf{Answer Specification} & \begin{tabular}[c]{@{}l@{}}Length of the answer, Existence of punctuation, Surface features (e.g URLs, Emojis,\\ Phones, Numbers, Not Persian(English or Persian typed in English or so on),\\or repetition of some letters(such as Helloooo, goood morniiing)
    \end{tabular} \\ 
    \hline
    \textbf{Question Answer Pair} & \begin{tabular}[c]{@{}l@{}}Different kinds of similarity between the question
    and its answers, Information\\achieve from the answers with attention to the questions
    \end{tabular}\\ 
    \hline
    \textbf{Metadata} & \begin{tabular}[c]{@{}l@{}}ID of the users (questioner or responder) whether they are the same (QUSERID\\and CUSERID) timestamp, question category, known number of like for an answer
    \end{tabular}\\ 
    \hline
    \end{tabular}
\caption{The categorized list of the features used by feature-based systems.}
\label{table:features categories}
\end{center}
\end{table*}
The features are classified into four groups that are shown in Table \ref{table:features categories}.
\begin{table}[!h]
\begin{center}
    \begin{tabular}{c|c|c|c}
    \hline
    \textbf{Answers} & \textbf{\#Good} & \textbf{\#Bad} & \textbf{\#potential} \\ 
    \hline
    21915 & 10467  & 10700 & 748 \\ 
    \hline
    100\%  & 47.8\%  & 48.7\%  & 3.14\% \\
    \hline
    \end{tabular}
\caption{Label distribution in PerCQA.}
\label{table:Distribution of Tags}
\end{center}
\end{table}

We manually extracted some simple features to perform feature engineering. This is performed to get a better understanding of the underlying latent relationships in data. The number of URLs, Emojis, Numbers, and Non-Persian words are reported in answers and help us to gain new findings about the data. Furthermore, it is important to consider the length of the answers when assessing classification quality. For instance, about 30\% of all answers that have less than 25 characters, as well as almost 20\% of answers that have emojis, belongs to the ``Bad'' class. A report on the distribution of dataset labels is shown in Table \ref{table:Distribution of Tags}. Although, because of our pre-trained model we do not require to perform feature extraction, we extracted a few simple features and depicted their correlation with the three classes. More details along are available in Appendix (feature engineering).

\section{Experiments}
\label{Section5}
 PerCQA offers us the opportunities to evaluate CQA systems on various tasks in Persian. In this section, we use PerCQA for answer selection, implement several baseline systems and evaluate and analyze their results.

\begin{table*}[ht]
\begin{center}
\begin{tabular}{|cccc|c|c|}
\hline
\multicolumn{4}{|c|}{\textbf{Word Embedding}} & \textbf{\begin{tabular}[c]{@{}c@{}}Various models for\\ Answer Selection\end{tabular}} & \textbf{F1-Score} \\ 
\hline
\multicolumn{1}{|c|}{\multirow{3}{*}{\begin{tabular}[c]{@{}c@{}}Non-\\ Comtextualized\end{tabular}}} & \multicolumn{3}{c|}{\multirow{2}{*}{Word2vec}} & PV-CNT & 41.03 \\ 
\cline{5-6} 
\multicolumn{1}{|c|}{}                                                                               & \multicolumn{3}{c|}{}                                                                                                                                                                                    & BiLSTM-attention                                                                       & 38.27             \\ \cline{2-6} 
\multicolumn{1}{|c|}{}                                                                               & \multicolumn{3}{c|}{FastText}                                                                                                                                                                            & RCNN                                                                                   & 39.56             \\ \hline
\multicolumn{1}{|c|}{\multirow{5}{*}{Contextualized}}                                                & \multicolumn{1}{c|}{Mono-Lingual}                   & \multicolumn{2}{c|}{\textbf{Pars-BERT}}                                                                                                            & CETE                                                                                   & \textbf{58.07}    \\ \cline{2-6} 
\multicolumn{1}{|c|}{}                                                                               & \multicolumn{1}{c|}{\multirow{4}{*}{Multi-Lingual}} & \multicolumn{1}{c|}{m-BERT}                          & \multirow{2}{*}{\begin{tabular}[c]{@{}c@{}}Fine-tuning\\ with PerCQA \end{tabular}}       & CETE                                                                                   & 50.71             \\ \cline{3-3} \cline{5-6} 
\multicolumn{1}{|c|}{}                                                                               & \multicolumn{1}{c|}{}                               & \multicolumn{1}{c|}{XLM-R}                           &                                                                                             & CETE                                                                                   & 54.13             \\ \cline{3-6} 
\multicolumn{1}{|c|}{}                                                                               & \multicolumn{1}{c|}{}                               & \multicolumn{1}{c|}{\multirow{2}{*}{\textbf{XLM-R}}} & \begin{tabular}[c]{@{}c@{}}Fine-tuning with\\ SemEvalCQA datasets\end{tabular}              & CETE                                                                                   & 52.48             \\ \cline{4-6} 
\multicolumn{1}{|c|}{}                                                                               & \multicolumn{1}{c|}{}                               & \multicolumn{1}{c|}{}                                & \begin{tabular}[c]{@{}c@{}}Fine-tuning with\\ SemEvalCQA datasets\\ and PerCQA\end{tabular} & CETE                                                                                   & \textbf{61.14}    \\ \hline
\end{tabular}
\caption{Quantitative evaluation results on PerCQA, ordered by macro-averaged F1. Each row corresponds to the result of an answer selection model using one of the pretrained word embeddings.}
\label{table:Our Experiments}
\end{center}
\end{table*}

\subsection{Baseline Methods}
For evaluating the dataset, we select the answer selection task and choose four baselines and apply them to PerCQA. We use one feature-based method~\cite{yang2015wikiqa},
and three deep-learning models. We use the implementation of these baselines which is described below:

\begin{itemize}
    \item \textbf{PV-Cnt}~\cite{yang2015wikiqa}\footnote{\url{https://gist.github.com/shagunsodhani/7cf3677ff2b0028a33e6702fbd260bc5}}: Word Count and Weighted Word Count are two word-matching features. Paragraph Vector~\cite{le2014distributed} is the cosine similarity score between the question vector and the sentence vector. They combined PV and word matching features by training a logistic regression classifier, referring as PV-Cnt.
    
    \item \textbf{BiLSTM-attention}~\cite{tan2015lstm}\footnote{\url{https://github.com/sachinbiradar9/Question-Answer-Selection}}: BiLSTM-attention is a basic framework to build the embeddings of questions and answers based on BiLSTM model. BiLSTM generates distributed representations by attention mechanism.
    \item \textbf{RCNN}~\cite{zhou2018recurrent}: Recurrent Convolutional Neural Network (RCNN) combines Convolutional Neural Network (CNN) with Recurrent Neural Network (RNN) to model the semantic relevance between questions and answers. 
    \item \textbf{CETE}~\cite{laskar2020contextualized}\footnote{\url{https://github.com/tahmedge/CETE-LREC}}: Contextualized Embeddings based Transformer Encoder (CETE) utilizes the pre-trained transformer encoder based models (BERT/RoBERTa) and integrates them for sentence similarity in the answer selection task. There are two approaches: i) feature-based approach, and ii) Fine-tuning-based approach.

\end{itemize}

\subsection{Word Embeddings}
We employ five pretrained word embeddings. Three are contextual, including ParsBERT, XLM-R and m-BERT.
Two others are non-contextual or static, including Word2vec and Fasttext.
For the contextual embeddings, we use the publicly available Pytorch models for each.
We train the non-contextual embeddings on a corpus made from Ninisite forums.
This corpus has about 2 billion tokens containing questions and answers from X threads.
The static embeddings such as Word2vec and FastText are employed on the proposed dataset and the dimension of word embedding is adopted w=200. 
Besides, we compare the performance of a mono-lingual model (ParsBERT) versus multi-lingual models (mBERT and XLM-R). 

\subsection{Results}
 Officially, in the previous work, macro-averaged F1 is used to benchmark the answer selection task. Therefore, we make the comparison based on this measurement. Table \ref{table:Our Experiments} demonstrates the results of these methods on the PerCQA dataset.
 We combine various word embedding (Word2vec, FastText, ParsBERT, mBERT, XLM-R) with diverse baselines methods (PV-CNT, BiLSTM-attention, RCNN, CETE-feature based approach).

It is evident that PV-CNT achieves better results than BiLSTM-attention while both of them utilize Word2vec as word embedding. Therefore, it we can conclude that lengthy and informal sentences may have been contributed to the deep learning-based method's low performance.

Models with contextual embeddings are significantly better than the ones with non-contextual embeddings. In PerCQA, there are often the lengthy questions, so considering the context for representing each word makes it possible to achieve a better result. 
We also compare the results of mBERT and XLM-R, the multi-lingual contextual embeddings, with mono-lingual embeddings of ParsBERT.
ParsBERT outperforms mBERT and XLM-R when training on the PerCQA data.
However, we observe that preraining XLM-R on SemEval English datasets is very effective as its macro F1 improves from 50.71 to 61.14.
This sets the best result on our PerCQA dataset better than ParsBERT's 58.07 F1, and shows that cross-lingual transfer from English datasets to our PerCQA dataset is possible. 
Even the zero-shot cross-lingual results of XLM-R are descent (52.48) compared to training on the same language data (54.13). 


\section{Conclusion}
\label{Section6}
We introduced PerCQA in the Persian Language for the task of answer selection in Community Question Answering (CQA). PerCQA contains 21,915 pairs of real questions and answers, which asked by a large number of users of various levels of literacy. We hope that PerCQA will promote the quality of Persian forums and enable further research in CQA tasks in Persian language. We plan to release a new version of PerCQA in the future, which we expect to include data to perform more CQA tasks in Persian. We also hope that our experimental results will provided practical baselines for further research.

\section{References}\label{reference}

\bibliographystyle{lrec2022-bib}
\bibliography{lrec2022-example}

\begin{thebibliography}{}

\bibitem[\protect\citename{{Agichtein} \bgroup et al.\egroup
  }2008]{agichtein2008finding}
{Agichtein}, E., {Castillo}, C., {Donato}, D., {Gionis}, A., and {Mishne}, G.
\newblock (2008).
\newblock Finding high-quality content in social media.
\newblock In {\em Proceedings of the 2008 International Conference on Web
  Search and Data Mining}, pages 183--194.

\bibitem[\protect\citename{Bojanowski \bgroup et al.\egroup
  }2017]{bojanowski2017enriching}
Bojanowski, P., Grave, E., Joulin, A., and Mikolov, T.
\newblock (2017).
\newblock Enriching word vectors with subword information.
\newblock {\em Transactions of the Association for Computational Linguistics},
  pages 135--146.

\bibitem[\protect\citename{Cer \bgroup et al.\egroup
  }2018]{DBLP:journals/corr/abs-1803-11175}
Cer, D., Yang, Y., Kong, S., Hua, N., Limtiaco, N., John, R.~S., Constant, N.,
  Guajardo{-}Cespedes, M., Yuan, S., Tar, C., Sung, Y., Strope, B., and
  Kurzweil, R.
\newblock (2018).
\newblock Universal sentence encoder.
\newblock {\em CoRR}.

\bibitem[\protect\citename{Conneau and Lample}2019]{conneau2019cross}
Conneau, A. and Lample, G.
\newblock (2019).
\newblock Cross-lingual language model pretraining.
\newblock {\em Advances in Neural Information Processing Systems}, pages
  7059--7069.

\bibitem[\protect\citename{Conneau \bgroup et al.\egroup
  }2020]{conneau-etal-2020-unsupervised}
Conneau, A., Khandelwal, K., Goyal, N., Chaudhary, V., Wenzek, G., Guzm{\'a}n,
  F., Grave, E., Ott, M., Zettlemoyer, L., and Stoyanov, V.
\newblock (2020).
\newblock Unsupervised cross-lingual representation learning at scale.
\newblock In {\em Proceedings of the 58th Annual Meeting of the Association for
  Computational Linguistics}, pages 8440--8451.

\bibitem[\protect\citename{Devlin \bgroup et al.\egroup
  }2019]{devlin-etal-2019-bert}
Devlin, J., Chang, M.-W., Lee, K., and Toutanova, K.
\newblock (2019).
\newblock {BERT}: Pre-training of deep bidirectional transformers for language
  understanding.
\newblock In {\em Proceedings of the 2019 Conference of the North {A}merican
  Chapter of the Association for Computational Linguistics: Human Language
  Technologies, Volume 1 (Long and Short Papers)}, pages 4171--4186.

\bibitem[\protect\citename{Farahani \bgroup et al.\egroup
  }2021]{farahani2021parsbert}
Farahani, M., Gharachorloo, M., Farahani, M., and Manthouri, M.
\newblock (2021).
\newblock Parsbert: Transformer-based model for persian language understanding.
\newblock {\em Neural Processing Letters}, pages 3831--3847.

\bibitem[\protect\citename{{Feng} \bgroup et al.\egroup
  }2015]{feng2015applying}
{Feng}, M., {Xiang}, B., {Glass}, M.~R., {Wang}, L., and {Zhou}, B.
\newblock (2015).
\newblock Applying deep learning to answer selection: A study and an open task.
\newblock In {\em 2015 IEEE Workshop on Automatic Speech Recognition and
  Understanding (ASRU)}, pages 813--820.

\bibitem[\protect\citename{Ghasemi \bgroup et al.\egroup
  }2021]{ghasemi2021user}
Ghasemi, N., Fatourechi, R., and Momtazi, S.
\newblock (2021).
\newblock User embedding for expert finding in community question answering.
\newblock {\em ACM Transactions on Knowledge Discovery from Data (TKDD)}, pages
  1--16.

\bibitem[\protect\citename{Grover and Leskovec}2016]{grover2016node2vec}
Grover, A. and Leskovec, J.
\newblock (2016).
\newblock node2vec: Scalable feature learning for networks.
\newblock In {\em Proceedings of the 22nd ACM SIGKDD international conference
  on Knowledge discovery and data mining}, pages 855--864.

\bibitem[\protect\citename{Hou \bgroup et al.\egroup
  }2015]{hou-etal-2015-hitsz}
Hou, Y., Tan, C., Wang, X., Zhang, Y., Xu, J., and Chen, Q.
\newblock (2015).
\newblock {HITSZ}-{ICRC}: Exploiting classification approach for answer
  selection in community question answering.
\newblock In {\em Proceedings of the 9th International Workshop on Semantic
  Evaluation ({S}em{E}val 2015)}, pages 196--202.

\bibitem[\protect\citename{{Huang} \bgroup et al.\egroup
  }2007]{huang2007extracting}
{Huang}, J., {Zhou}, M., and {Yang}, D.
\newblock (2007).
\newblock Extracting chatbot knowledge from online discussion forums.
\newblock In {\em IJCAI'07 Proceedings of the 20th international joint
  conference on Artifical intelligence}, pages 423--428.

\bibitem[\protect\citename{Joulin \bgroup et al.\egroup
  }2016]{joulin2016fasttext}
Joulin, A., Grave, E., Bojanowski, P., Douze, M., J{\'e}gou, H., and Mikolov,
  T.
\newblock (2016).
\newblock Fasttext. zip: Compressing text classification models.
\newblock {\em arXiv preprint arXiv:1612.03651}.

\bibitem[\protect\citename{Kunneman \bgroup et al.\egroup
  }2019]{kunneman-etal-2019-question}
Kunneman, F., Ferreira, T.~C., Krahmer, E., and van~den Bosch, A.
\newblock (2019).
\newblock Question similarity in community question answering: A systematic
  exploration of preprocessing methods and models.
\newblock In {\em Proceedings of the International Conference on Recent
  Advances in Natural Language Processing (RANLP 2019)}, pages 593--601.

\bibitem[\protect\citename{{Laskar} \bgroup et al.\egroup
  }2020]{laskar2020contextualized}
{Laskar}, M. T.~R., {Huang}, J.~X., and {Hoque}, E.
\newblock (2020).
\newblock Contextualized embeddings based transformer encoder for sentence
  similarity modeling in answer selection task.
\newblock In {\em Proceedings of The 12th Language Resources and Evaluation
  Conference}, pages 5505--5514.

\bibitem[\protect\citename{Le and Mikolov}2014]{le2014distributed}
Le, Q. and Mikolov, T.
\newblock (2014).
\newblock Distributed representations of sentences and documents.
\newblock In {\em International conference on machine learning}, pages
  1188--1196.

\bibitem[\protect\citename{Liu \bgroup et al.\egroup }2019]{liu2019roberta}
Liu, Y., Ott, M., Goyal, N., Du, J., Joshi, M., Chen, D., Levy, O., Lewis, M.,
  Zettlemoyer, L., and Stoyanov, V.
\newblock (2019).
\newblock Roberta: A robustly optimized bert pretraining approach.
\newblock {\em arXiv e-prints}, pages arXiv--1907.

\bibitem[\protect\citename{{Mihaylov} and {Nakov}}2016]{mihaylov2016semanticz}
{Mihaylov}, T. and {Nakov}, P.
\newblock (2016).
\newblock Semanticz at semeval-2016 task 3: Ranking relevant answers in
  community question answering using semantic similarity based on fine-tuned
  word embeddings.
\newblock In {\em Proceedings of the 10th International Workshop on Semantic
  Evaluation (SemEval-2016)}, pages 879--886.

\bibitem[\protect\citename{Mikolov \bgroup et al.\egroup
  }2013]{DBLP:journals/corr/abs-1301-3781}
Mikolov, T., Chen, K., Corrado, G., and Dean, J.
\newblock (2013).
\newblock Efficient estimation of word representations in vector space.
\newblock In {\em 1st International Conference on Learning Representations,
  {ICLR} 2013, Scottsdale, Arizona, USA, May 2-4, 2013, Workshop Track
  Proceedings}.

\bibitem[\protect\citename{Nakov \bgroup et al.\egroup
  }2015]{nakov-etal-2015-semeval}
Nakov, P., M{\`a}rquez, L., Magdy, W., Moschitti, A., Glass, J., and Randeree,
  B.
\newblock (2015).
\newblock {S}em{E}val-2015 task 3: Answer selection in community question
  answering.
\newblock In {\em Proceedings of the 9th International Workshop on Semantic
  Evaluation ({S}em{E}val 2015)}, pages 269--281.

\bibitem[\protect\citename{Nakov \bgroup et al.\egroup
  }2016a]{nakov-etal-2016-semeval-2016}
Nakov, P., M{\`a}rquez, L., Moschitti, A., Magdy, W., Mubarak, H., Freihat,
  A.~A., Glass, J., and Randeree, B.
\newblock (2016a).
\newblock {S}em{E}val-2016 task 3: Community question answering.
\newblock In {\em Proceedings of the 10th International Workshop on Semantic
  Evaluation ({S}em{E}val-2016)}, pages 525--545.

\bibitem[\protect\citename{{Nakov} \bgroup et al.\egroup }2016b]{nakov2016it}
{Nakov}, P., {Màrquez}, L., and {Guzmán}, F.
\newblock (2016b).
\newblock It takes three to tango: Triangulation approach to answer ranking in
  community question answering.
\newblock In {\em Proceedings of the 2016 Conference on Empirical Methods in
  Natural Language Processing}, pages 1586--1597.

\bibitem[\protect\citename{Nakov \bgroup et al.\egroup
  }2017]{nakov-etal-2017-semeval}
Nakov, P., Hoogeveen, D., M{\`a}rquez, L., Moschitti, A., Mubarak, H., Baldwin,
  T., and Verspoor, K.
\newblock (2017).
\newblock {S}em{E}val-2017 task 3: Community question answering.
\newblock In {\em Proceedings of the 11th International Workshop on Semantic
  Evaluation ({S}em{E}val-2017)}, pages 27--48.

\bibitem[\protect\citename{Nicosia \bgroup et al.\egroup
  }2015]{nicosia2015qcri}
Nicosia, M., Filice, S., Barr{\'o}n-Cedeno, A., Saleh, I., Mubarak, H., Gao,
  W., Nakov, P., MARTINO, G. D.~S., Moschitti, A., Darwish, K., et~al.
\newblock (2015).
\newblock Qcri: Answer selection for community question answering-experiment
  for arabic and english.
\newblock In {\em Proceedings of the 9th International Workshop on Semantic
  Evaluation, SemEval@NAACL-HLT 2015, Denver, Colorado, USA, June 4-5, 2015}.

\bibitem[\protect\citename{{Omari} \bgroup et al.\egroup
  }2016]{omari2016novelty}
{Omari}, A., {Carmel}, D., {Rokhlenko}, O., and {Szpektor}, I.
\newblock (2016).
\newblock Novelty based ranking of human answers for community questions.
\newblock In {\em Proceedings of the 39th International ACM SIGIR conference on
  Research and Development in Information Retrieval}, pages 215--224.

\bibitem[\protect\citename{{Othman} \bgroup et al.\egroup }2017]{othman2017a}
{Othman}, N., {Faiz}, R., and {Smaili}, K.
\newblock (2017).
\newblock A word embedding based method for question retrieval in community
  question answering.
\newblock In {\em ICNLSSP 2017 - International Conference on Natural Language,
  Signal and Speech Processing}.

\bibitem[\protect\citename{{Othman} \bgroup et al.\egroup
  }2019]{othman2019manhattan}
{Othman}, N., {Faiz}, R., and {Smaïli}, K.
\newblock (2019).
\newblock Manhattan siamese lstm for question retrieval in community question
  answering.
\newblock {\em OTM Confederated International Conferences On the Move to
  Meaningful Internet Systems}, pages 661--677.

\bibitem[\protect\citename{Pennington \bgroup et al.\egroup
  }2014]{pennington-etal-2014-glove}
Pennington, J., Socher, R., and Manning, C.
\newblock (2014).
\newblock {G}lo{V}e: Global vectors for word representation.
\newblock In {\em Proceedings of the 2014 Conference on Empirical Methods in
  Natural Language Processing ({EMNLP})}, pages 1532--1543.

\bibitem[\protect\citename{Peters \bgroup et al.\egroup }2018]{peters2018deep}
Peters, M., Neumann, M., Iyyer, M., Gardner, M., Clark, C., Lee, K., and
  Zettlemoyer, L.
\newblock (2018).
\newblock Deep contextualized word representations.
\newblock In {\em Proceedings of the 2018 Conference of the North American
  Chapter of the Association for Computational Linguistics: Human Language
  Technologies, Volume 1 (Long Papers)}, pages 2227--2237.

\bibitem[\protect\citename{{Qiu} and {Huang}}2015]{qiu2015convolutional}
{Qiu}, X. and {Huang}, X.
\newblock (2015).
\newblock Convolutional neural tensor network architecture for community-based
  question answering.
\newblock In {\em IJCAI'15 Proceedings of the 24th International Conference on
  Artificial Intelligence}, pages 1305--1311.

\bibitem[\protect\citename{Qiu \bgroup et al.\egroup }2018]{qiu2018network}
Qiu, J., Dong, Y., Ma, H., Li, J., Wang, K., and Tang, J.
\newblock (2018).
\newblock Network embedding as matrix factorization: Unifying deepwalk, line,
  pte, and node2vec.
\newblock In {\em Proceedings of the eleventh ACM international conference on
  web search and data mining}, pages 459--467.

\bibitem[\protect\citename{Radford and Narasimhan}2018]{Radford2018ImprovingLU}
Radford, A. and Narasimhan, K.
\newblock (2018).
\newblock Improving language understanding by generative pre-training.

\bibitem[\protect\citename{Tan \bgroup et al.\egroup }2015]{tan2015lstm}
Tan, M., dos Santos, C., Xiang, B., and Zhou, B.
\newblock (2015).
\newblock Lstm-based deep learning models for non-factoid answer selection.
\newblock {\em arXiv e-prints}, pages arXiv--1511.

\bibitem[\protect\citename{Tran \bgroup et al.\egroup
  }2015]{tran-etal-2015-jaist}
Tran, Q.~H., Tran, V.~D., Vu, T.~T., Nguyen, M.~L., and Pham, S.~B.
\newblock (2015).
\newblock {JAIST}: Combining multiple features for answer selection in
  community question answering.
\newblock In {\em Proceedings of the 9th International Workshop on Semantic
  Evaluation ({S}em{E}val 2015)}, pages 215--219.

\bibitem[\protect\citename{{Vaswani} \bgroup et al.\egroup
  }2017]{vaswani2017attention}
{Vaswani}, A., {Shazeer}, N., {Parmar}, N., {Uszkoreit}, J., {Jones}, L.,
  {Gomez}, A.~N., {Kaiser}, L., and {Polosukhin}, I.
\newblock (2017).
\newblock Attention is all you need.
\newblock In {\em Proceedings of the 31st International Conference on Neural
  Information Processing Systems}, pages 5998--6008.

\bibitem[\protect\citename{{Wang} \bgroup et al.\egroup }2007]{wang2007what}
{Wang}, M., {Smith}, N.~A., and {Mitamura}, T.
\newblock (2007).
\newblock What is the jeopardy model? a quasi-synchronous grammar for qa.
\newblock In {\em Proceedings of the 2007 Joint Conference on Empirical Methods
  in Natural Language Processing and Computational Natural Language Learning
  (EMNLP-CoNLL)}, pages 22--32.

\bibitem[\protect\citename{Wang \bgroup et al.\egroup
  }2020]{10.1145/3397271.3401143}
Wang, Z., Fan, Y., Guo, J., Yang, L., Zhang, R., Lan, Y., Cheng, X., Jiang, H.,
  and Wang, X.
\newblock (2020).
\newblock Match²: A matching over matching model for similar question
  identification.
\newblock In {\em Proceedings of the 43rd International ACM SIGIR Conference on
  Research and Development in Information Retrieval}, page 559–568.

\bibitem[\protect\citename{{Wen} \bgroup et al.\egroup }2019]{wen2019joint}
{Wen}, J., {Tu}, H., {Cheng}, X., {Xie}, R., and {Yin}, W.
\newblock (2019).
\newblock Joint modeling of users, questions and answers for answer selection
  in cqa.
\newblock {\em Expert Systems With Applications}, pages 563--572.

\bibitem[\protect\citename{{Xiang} \bgroup et al.\egroup
  }2016]{xiang2016incorporating}
{Xiang}, Y., {Zhou}, X., {Chen}, Q., {Zheng}, Z., {Tang}, B., {Wang}, X., and
  {Qin}, Y.
\newblock (2016).
\newblock Incorporating label dependency for answer quality tagging in
  community question answering via cnn-lstm-crf.
\newblock In {\em Proceedings of COLING 2016, the 26th International Conference
  on Computational Linguistics: Technical Papers}, pages 1231--1241.

\bibitem[\protect\citename{{Xiang} \bgroup et al.\egroup
  }2017]{xiang2017answer}
{Xiang}, Y., {Chen}, Q., {Wang}, X., and {Qin}, Y.
\newblock (2017).
\newblock Answer selection in community question answering via attentive neural
  networks.
\newblock {\em IEEE Signal Processing Letters}, pages 505--509.

\bibitem[\protect\citename{{Yang} \bgroup et al.\egroup }2015]{yang2015wikiqa}
{Yang}, Y., tau {Yih}, W., and {Meek}, C.
\newblock (2015).
\newblock Wikiqa: A challenge dataset for open-domain question answering.
\newblock In {\em Proceedings of the 2015 Conference on Empirical Methods in
  Natural Language Processing}, pages 2013--2018.

\bibitem[\protect\citename{{Yang} \bgroup et al.\egroup
  }2019]{yang2019advanced}
{Yang}, M., {Tu}, W., {Qu}, Q., {Zhou}, W., {Liu}, Q., and {Zhu}, J.
\newblock (2019).
\newblock Advanced community question answering by leveraging external
  knowledge and multi-task learning.
\newblock {\em Knowledge Based Systems}, pages 106--119.

\bibitem[\protect\citename{Zhong \bgroup et al.\egroup }2020]{zhong2020jec}
Zhong, H., Xiao, C., Tu, C., Zhang, T., Liu, Z., and Sun, M.
\newblock (2020).
\newblock Jec-qa: A legal-domain question answering dataset.
\newblock In {\em Proceedings of the AAAI Conference on Artificial
  Intelligence}, pages 9701--9708.

\bibitem[\protect\citename{{Zhou} \bgroup et al.\egroup
  }2015]{zhou2015learning}
{Zhou}, G., {He}, T., {Zhao}, J., and {Hu}, P.
\newblock (2015).
\newblock Learning continuous word embedding with metadata for question
  retrieval in community question answering.
\newblock In {\em Proceedings of the 53rd Annual Meeting of the Association for
  Computational Linguistics and the 7th International Joint Conference on
  Natural Language Processing (Volume 1: Long Papers)}, pages 250--259.

\bibitem[\protect\citename{{Zhou} \bgroup et al.\egroup
  }2018]{zhou2018recurrent}
{Zhou}, X., {Hu}, B., {Chen}, Q., and {Wang}, X.
\newblock (2018).
\newblock Recurrent convolutional neural network for answer selection in
  community question answering.
\newblock {\em Neurocomputing}, pages 8--18.

\end{thebibliography}

\label{lr:ref}
\bibliographystylelanguageresource{lrec2022-bib}

\section{Appendix}
\subsection{Details about PerCQA}
\begin{figure*}[!h]
\begin{center}
\includegraphics[scale=0.5]{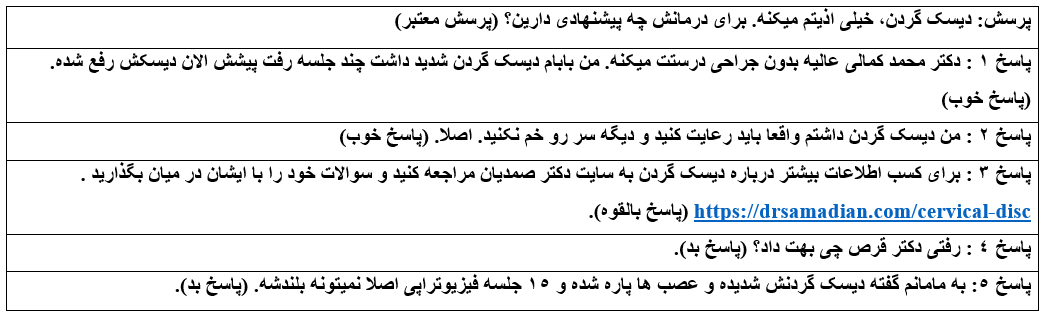}

\caption{A question and some of its answers in PerCQA.} 
\label{fig.1}
\end{center}
\end{figure*}

\begin{figure*}[!h]
\begin{center}
\includegraphics[scale=0.5]{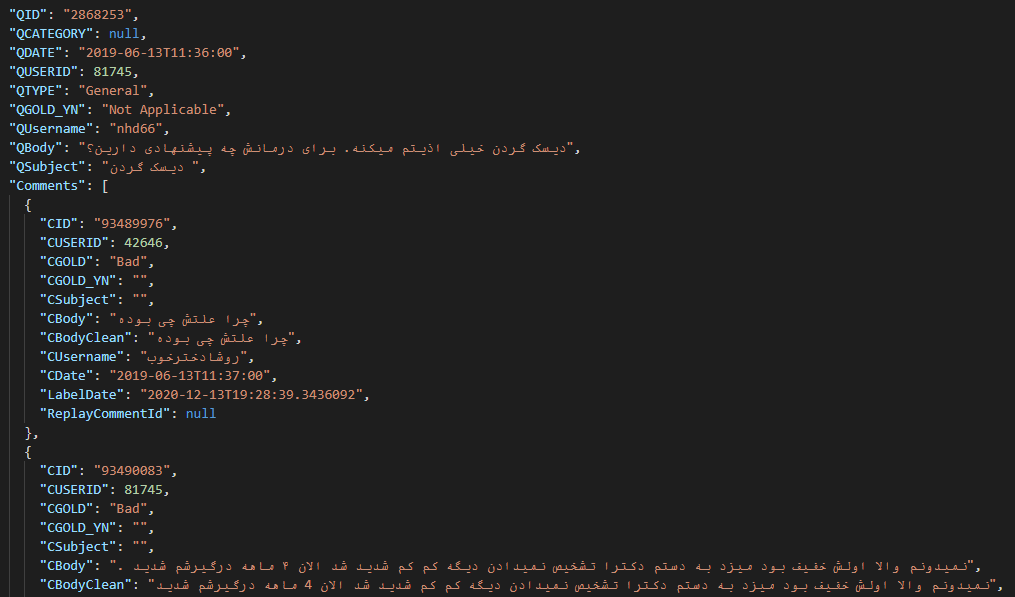}

\caption{A sample of annotated thread from PerCQA (JSON-formated).} 
\label{fig.2}
\end{center}
\end{figure*}

Figure \ref{fig.1} indicates the original version of the question and some of its answers in the proposed dataset which the translation of this sample is shown in Table \ref{table:Sample question and its answers}. Figure \ref{fig.2} reveals JSON-formated of this question in PerCQA. There are characteristics for the questions and the answers that do not have values that will be relevant to future work in this research. For example QGOLD\_YN. 
There is  one of the latest questions on the subject of ``Cervical Disc'' in this link\footnote{\url{https://www.ninisite.com/discussion/topic/6365515}}. As you can see, as mentioned before, it is asked and answered in a very informal way.

\subsection{Data labelling Quality}
Cohen's kappa coefficient is a statistic that is used to measure inter-rater reliability for categorical items. The judges' harmony levels with the final labels and the number of agreements on the matrix's main diagonal is shown in Table \ref{table:number of agreements}. The observed proportionate agreement is approximately 0.8, which Table \ref{table:Cohen’s kappa criteria} reveals it.
\begin{table}[!h]
\begin{center}
    \begin{tabularx}{\columnwidth}{c|c|c|c|c}
    \hline
    \textbf{Labels}    & \textbf{Good} & \textbf{Bad} & \textbf{Potential} & \textbf{Total} \\ 
    \hline
    \textbf{Good}      & 4195  & 59  & 128  & 4382 \\ 
    \hline
    \textbf{Bad}       & 412 & 4216  & 368  & 4996 \\ 
    \hline
    \textbf{Potential} & 79  & 63    & 384  & 526  \\ 
    \hline
    \textbf{Total}     & 4686 & 4358 & 880  & 9924 \\ 
    \hline
    \end{tabularx}
\caption{The number of agreements and disagreements in the labeling process to calculate the kappa criterion.}
\label{table:number of agreements}
\end{center}
\end{table}
\begin{table}[!h]
\begin{center}
    \begin{tabularx}{\columnwidth}{c|c}
    \hline
    \begin{tabular}[c]{@{}c@{}}Unweighted kappa \end{tabular} & \begin{tabular}[c]{@{}c@{}}Kappa with linear weighted
    \end{tabular} \\ 
    \hline  0.802 & 0.7892   \\ 
    \hline
    \end{tabularx}
\caption{Cohen’s kappa criteria.}
\label{table:Cohen’s kappa criteria}
\end{center}
\end{table}

\subsection{Feature Engineering}
\begin{figure}[!h]
\begin{center}
\includegraphics[scale=0.5]{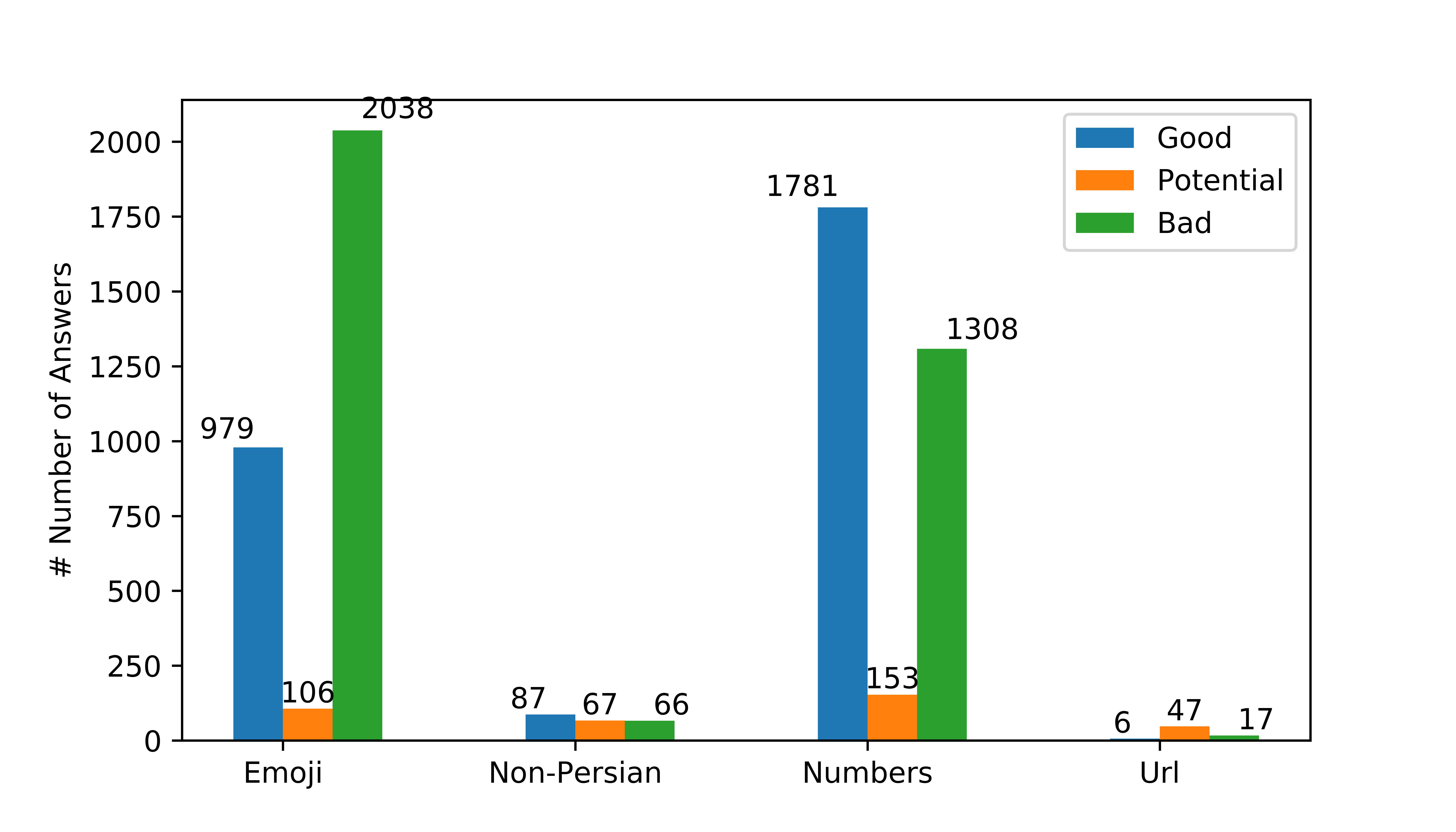} 

\caption{The number of Answers that have some features }
\label{fig.3}
\end{center}
\end{figure}
\begin{figure}[!h]
\begin{center}
\includegraphics[scale=0.5]{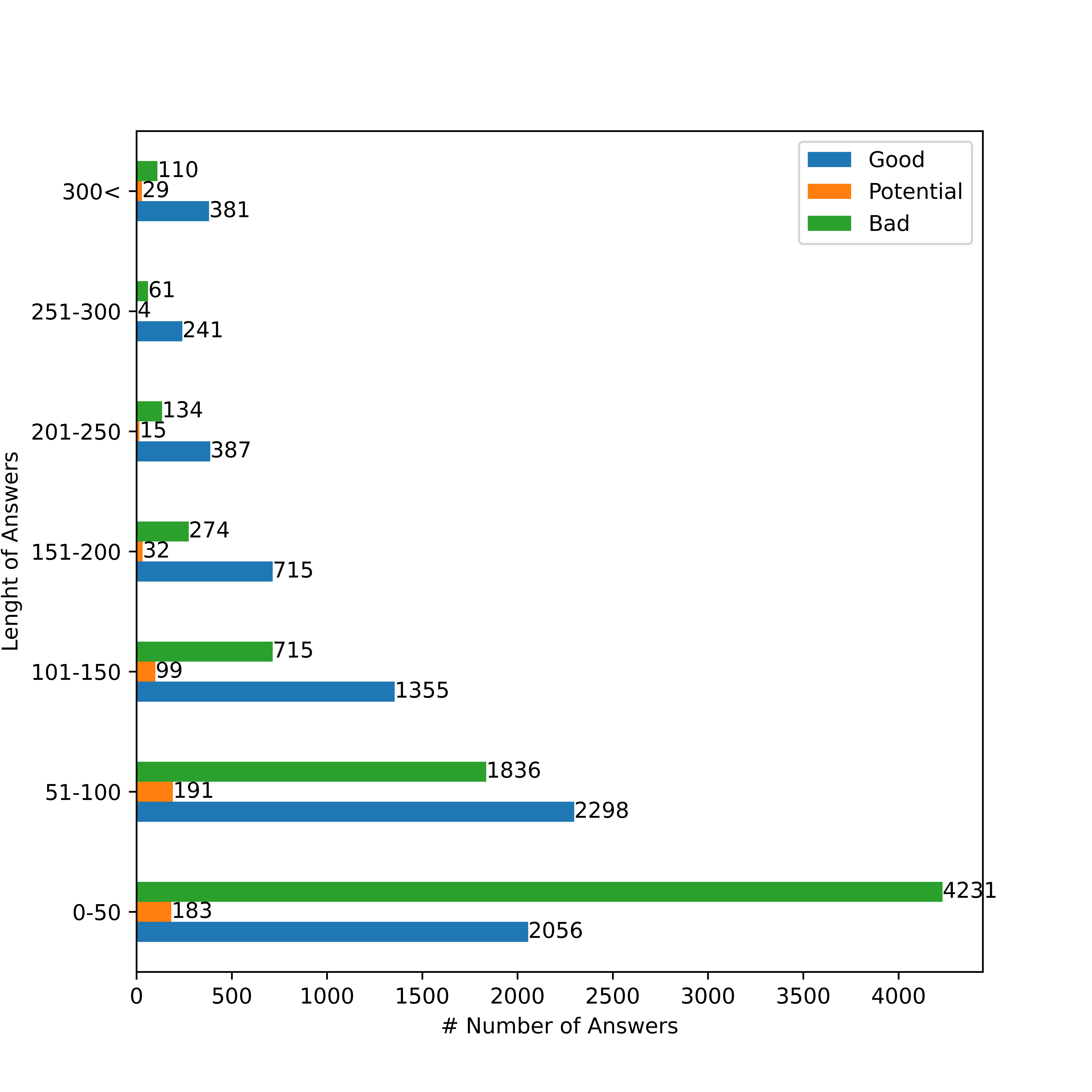}

\caption{The distribution of answer labels based on their length}
\label{fig.4}
\end{center}
\end{figure}
Feature engineering involves extracting features (characteristics, properties, attributes) from raw data and it is extremely important in Machine Learning (ML) and helps data scientists evaluate trade-off decisions regarding the impact of ML models. Considering that the importance of features varies, we extract some obvious features that are more common in a particular class than others. 

For example, there are more stickers in the answers that have a ``Bad'' labels. The number of ``non-Persian'' items in ``Good'' labels is higher than others, because the names of drugs, cosmetics, home appliances or dowry brands are in English. In addition, phone numbers and digits exist in the more answers with ``Good'' labels significantly and one-Tenth of the ``Potential'' tags have URLs. Figure \ref{fig.3} reveals these features.

 As referred earlier, another feature that affects the quality of classification is the length of answers. The number of answers in different lengths according to their tags are shown in figure \ref{fig.4}. As it can be seen, the answers with fewer than 25 characters are more likely to be labelled ``Bad'' and decrease sharply with increasing length.

\end{document}